\documentclass{article}

\usepackage[utf8]{inputenc} % allow utf-8 input
\usepackage[T1]{fontenc}    % use 8-bit T1 fonts
\usepackage{hyperref}       % hyperlinks
\usepackage{url}            % simple URL typesetting
\usepackage{booktabs}       % professional-quality tables
\usepackage{amsfonts}       % blackboard math symbols
\usepackage{nicefrac}       % compact symbols for 1/2, etc.
\usepackage{microtype}      % microtypography

\usepackage{bm}
\usepackage{bbm}
\usepackage{amsmath}
\usepackage{graphicx}

\newcommand{\by}{{\bf y}}
\newcommand{\bY}{{\bf Y}}
\newcommand{\bX}{{\bf X}}
\newcommand{\bff}{{\bf f}}
\newcommand{\bx}{{\bf x}}

\newcommand{\bo}{{\bm 1}}
\newcommand{\norm}{{\mathrm{N}}}
\newcommand{\bz}{{\bf z}}

\usepackage[left=3cm,right=3cm,top=3cm,bottom=3cm]{geometry}
\usepackage[backend=bibtex,style=nature]{biblatex}
\bibliography{nips_phenotime}
\title{Stratification of patient trajectories using covariate latent variable models}

\author{
  Kieran R. Campbell \& Christopher Yau \\ % \thanks{Use footnote for providing further
  Wellcome Trust Centre for Human Genetics \\
  University of Oxford \\
  \texttt{kieran.campbell@sjc.ox.ac.uk} \\
}

\begin{document}

\maketitle

\begin{abstract}

Standard models assign disease progression to discrete categories or \emph{stages} based on well-characterized clinical markers. However, such a system is potentially at odds with our understanding of the underlying biology, which in highly complex systems may support a (near-)continuous evolution of disease from inception to terminal state. To learn such a continuous disease score one could infer a latent variable from dynamic ``omics'' data such as RNA-seq that correlates with an outcome of interest such as survival time. However, such analyses may be confounded by additional data such as clinical covariates measured in electronic health records (EHRs). As a solution to this we introduce \emph{covariate latent variable models}, a novel type of latent variable model that learns a low-dimensional data representation in the presence of two (asymmetric) views of the same data source. We apply our model to TCGA colorectal cancer RNA-seq data and demonstrate how incorporating microsatellite-instability (MSI) status as an external covariate allows us to identify genes that stratify patients on an immune-response trajectory. Finally, we propose an extension termed \emph{Covariate Gaussian Process Latent Variable Models} for learning nonparametric, nonlinear representations. An \texttt{R} package implementing variational inference for covariate latent variable models is available at \url{http://github.com/kieranrcampbell/clvm}.

\end{abstract}

\section{Introduction}

There exists a set of physical processes with an assumed temporal component but where precise measurement of times associated with events is precluded or impossible. % KC - note sure about this sentence
Such ideas have recently flourished in the field of single-cell genomics, where cells will undergo some dynamic process such as differentiation but in which the destructive measurement of gene expression precludes physical measurement of the progression itself. Consequently, the progression is artificially inferred from the measured expression data as a \emph{pseudotime} (e.g. \cite{trapnell2014dynamics,delorean}), which in a statistical sense is akin to inference of a one-dimensional latent variable model.

\begin{figure}[h]
  \centering
  \includegraphics[width=.7\textwidth]{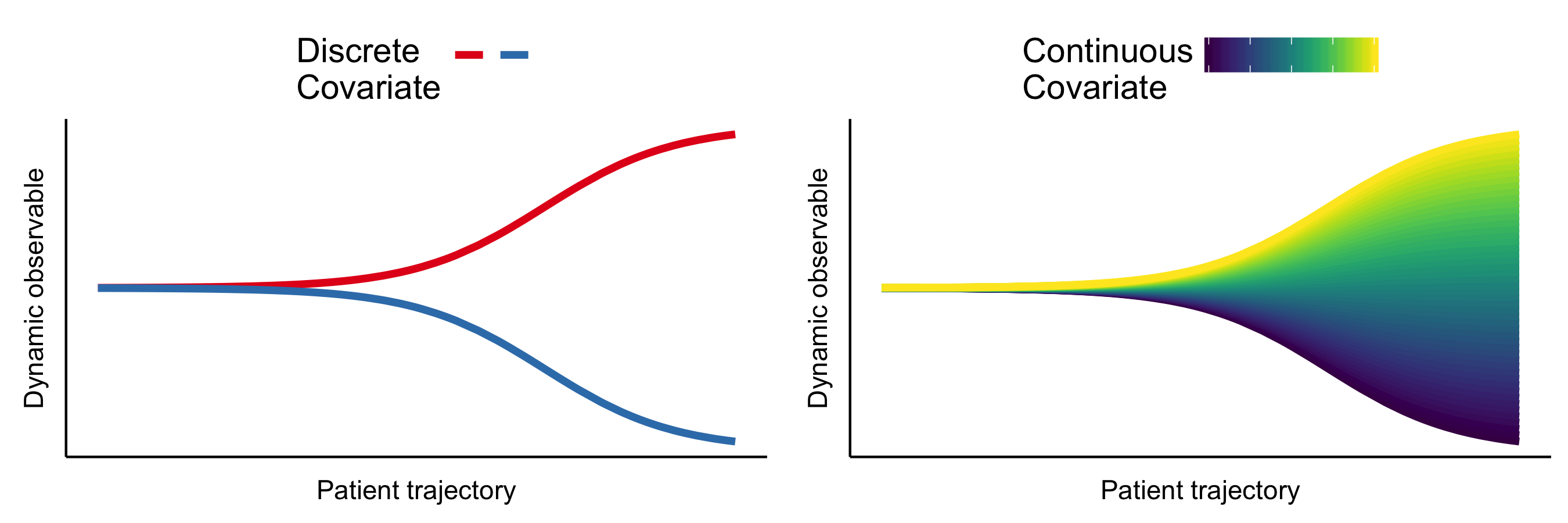}
  \caption{C-LVMs can be applied to infer trajectories in the presence of discrete covariates (left) or continuous covariates (right).}
 \label{fig:phenotime}
\end{figure}

This situation also arises in the case of disease staging and survival analysis such as when a patient presents to a clinic with a disease of unknown progression. Typically, the patient will be assigned a discrete disease \emph{stage} after possibly invasive tests and/or surgery. The discrete nature of such staging is at odds with accepted knowledge of the underlying biology, which is consistent with a more continuous evolution of disease progression such as gradual changes in gene expression. Furthermore, such evolution is confounded by underlying population heterogeneity, where the evolution of molecular features along the trajectory may differ depending on external patient phenotypes, such as age and sex or molecular phenotypes such as mutations of a particular gene (figure \ref{fig:phenotime}).

As a proof-of-concept solution to such issues we propose \emph{Covariate Latent Variable Models} (C-LVMs), a novel type of latent variable model similar to factor analysis in which the evolution of various dynamic genomic observables (such as gene expression) is allowed to vary according to a secondary set of covariates (such as mutation status). Such latent variable models combine two \emph{views} of the same data but where the relationship between each view and the latent variables is asymmetric. Formulated as a Bayesian hierarchical model we are further able to extract interactions between the patient trajectory and covariates, simultaneously providing insight into the underlying biology. We apply our model to RNA-seq data for the TCGA colorectal cancer dataset using microsatellite instability status as a covariate and extract a trajectory consistent with known markers of colorectal cancer prognosis. Finally, we propose a nonlinear, nonparametric extention and discuss the relationship to Gaussian Process Latent Variable Models.

\section{Methods}

\subsection{Model}

We begin with an $N \times G$ data matrix $\bY$ where $y_{ig}$ denotes the $i^{th}$ entry in the $g^{th}$ column for $i \in 1, \ldots, N$ samples and $g \in 1, \ldots, G$ features. Such a matrix would correspond to the measurement of a dynamic molecular process that we might reasonably expect to show continuous evolution as a disease progresses such as gene expression corresponding to a particular pathway. It is then trivial to learn a one-dimensional linear embedding that would be our ``best guess'' of such progression via a factor analysis model:
\begin{equation}
	y_{ig} = c_g z_i + \epsilon_{ig},  \; \epsilon_{ig} \sim \mathrm{N}(0,\tau_g^{-1})
\end{equation}
where $z_i$ is the latent measure of progression for sample $i$ and $c_g$ is the factor loading for feature $g$ which essentially describes the evolution of $g$ along the patient trajectory.

However, it is conceivable that the evolution of feature $g$ along the trajectory is not identical for all samples but is instead affected by a set of external covariates. Such covariates may correspond to patient phenotypes such as age or sex, EHR entries such as blood pressure or additional molecular data such as the mutation status of a particular gene. Note that we expect such features to be ``static'' and not necessarily correlate with the trajectory itself. 

Introducing the $N \times P$ covariate matrix $\bX$ with the entry in the $i^{th}$ row and $p^{th}$ column given by $x_{ip}$, we allow such measurements to perturb the factor loading matrix
\begin{equation}
	c_g \rightarrow \lambda_{ig} = c_g + \sum_{p=1}^P \beta_{pg} x_{ip}
\end{equation}
where $\beta_{pg}$ quantifies the effect of covariate $p$ on the evolution of feature $g$. Despite $\bY$ being column-centred we need to reintroduce gene and covariate specific intercepts to satisfy the model assumptions, giving a generative model of the form
\begin{equation}
	y_{ig} = \eta_g + \sum_{p=1}^P \alpha_{pg} x_{ip} + \left( c_g + \sum_{p=1}^P \beta_{pg} x_{ip}\right) z_i + \epsilon_{ig}, \; \;  \epsilon_{ig}\sim \mathrm{N}(0,\tau_g^{-1}) 
\end{equation}

% KC - taken out from before to make shorter
%\begin{equation}
%\begin{aligned}
%\chi_{pg} & \sim \mathrm{Gamma}(a_\beta, b_\beta) \\
%\beta_{pg} & \sim \mathrm{N}(0, \tau_g^{-1} \chi_{pg}^{-1})
%\end{aligned}
%\end{equation}

Our goal is inference of $z_i$ that encodes disease progression along with $\beta_{pg}$ which is informative of novel interactions between continuous trajectories and external covariates. Consequently we place a sparse Bayesian prior on $\beta_{pg}$ of the form $\beta_{pg} \sim \mathrm{N}(0, \chi_{pg}^{-1})$ where the posterior of $\chi_{pg}$ is informative of the model's belief that $\beta_{pg}$ is non-zero. The complete generative model is therefore given by

\begin{equation}
\begin{aligned}
\alpha_{pg} & \sim \norm(0, \tau_\alpha^{-1}) \\
c_g & \sim \norm(0, \tau_c^{-1}) \\
z_i & \sim \norm(q_i, \tau_q^{-1}) \\
\beta_{pg} & \sim \norm(0, \chi_{pg}^{-1}) \\
\chi_{pg}^{-1} & \sim \text{Gamma}(a_\beta, b_\beta) \\
\tau_{g}^{-1} & \sim \text{Gamma}(a, b) \\
\mu_{g} & \sim \norm(0, \tau_\mu^{-1}) \\
\epsilon_{ig} & \sim \norm(0, \tau_g^{-1}) \\
y_{ig} & = \mu_g + \sum_p \alpha_{pg} x_{ip} + \left( c_g + \sum_p \beta_{pg} x_{ip} \right) z_i + \epsilon_{ig}
\end{aligned}
\end{equation}

where $\tau_\alpha$, $\tau_c$, $a$, $b$, $a_\beta$, $b_\beta$, $\tau_q$ are fixed hyperparameters and $q_i$ encodes prior information about $z_i$ if available but typically $q_i = 0 \; \forall i$ in the uninformative case.

To understand this model it helps to consider the distribution of $\bY$ marginalised over the mapping $\{ c_g, \alpha_{pg}, \beta_{pg}\} \; \forall \; p,g$ with priors $c_g \sim \mathrm{N}(0, \tau_c^{-1})$ and $\alpha_{pg} \sim \mathrm{N}(0, \tau_\alpha^{-1})$. If $\by_g$ denotes the column vectors of $\bY$ and similarly $\bx_p$ for $\bX$, $[\bz]_i = z_i$, $\bo_N$ is the column vector of ones and $\odot$ denotes the element-wise product, then

\begin{equation}
	p(\by_g | \bX, \bz, \eta_g, \tau_g, \tau_c, \tau_\alpha, \chi_{pg}) \sim
\mathrm{N}\left(
\eta_g \bo_N, \bm \Sigma^{(g)}
\right)
\end{equation}

where

\begin{equation}
\bm \Sigma^{(g)} = 
\tau_g^{-1} \bo_N +
\tau_\alpha^{-1} \bX \bX^T +
\tau_c^{-1} \bz \bz^T +
 \sum_p  \chi_{pg}^{-1} (\bx_p \odot \bz) (\bx_p \odot \bz)^T.
\end{equation}

We therefore see that the addition of the covariates adds extra terms to the covariance matrix corresponding to \emph{perturbations} of the latent variables with the covariates. Consequently, the scale on which $\bx_p$ is defined needs carefully calibrated. Furthermore, it is possible to extend the latent variable matrix to have dimension larger than 1 giving a novel dimensionality reduction technique for visualisation, though additional rotation issues arise.

\subsection{Inference}

We perform co-ordinate ascent mean field variational inference (see e.g. \cite{blei2016variational}) with an approximating distribution of the form

\begin{equation}
\begin{aligned}
& q\left( 
\{ z_i \}_{i=1}^N, 
\{ \mu_g \}_{g=1}^G, 
\{ \tau_g \}_{g=1}^G,
\{ c_g \}_{g=1}^G,
\{ \alpha_{pg} \}_{g=1,p=1}^{G,P}
\{ \beta_{pg} \}_{g=1,p=1}^{G,P}
\{ \chi_{pg} \}_{g=1,p=1}^{G,P} 
\right) \\
& = \prod_{i=1}^N \underbrace{q_z(z_i)}_{\text{Normal}}
\prod_{g=1}^G \underbrace{q_\mu(\mu_g)}_{\text{Normal}} 
\underbrace{q_\tau(\tau_g)}_{\text{Gamma}} \underbrace{q_c(c_g)}_{\text{Normal}}
\prod_{p=1}^P \underbrace{q_\alpha (\alpha_{pg})}_{\text{Normal}} 
\underbrace{q_\beta(\beta_{pg})}_{\text{Normal}} \underbrace{q_\chi (\chi_{pg})}_{\text{Gamma}}
\end{aligned}
\end{equation}

Due to the model's conjugacy the optimal update for each parameter $\theta_j$ given all other parameters $\bm \theta_{-j}$ can easily be computed via

\begin{equation}
q^*_j(\theta_j) \propto \exp \left\{  \mathbf{E}_{-j} \left[ \log p(\theta_j | \bm \theta_{-j}, \bX, \bY) \right]
 \right\}
\end{equation}

where the expectation is taken with respect to the variational density over $\bm \theta_{-j}$.

\subsection{Identifying significant interactions} \label{sec:significant}

For each gene $g$ and covariate $p$ we have $\beta_{pg}$ that encodes the effect of $p$ on the evolution of $g$ along the trajectory $\bz$. We would like to identify interesting or \emph{significant} interactions for further analysis and follow up. 

The variational approximation for $\beta_{pg}$ is given by

\begin{equation}
\begin{aligned}
q_{\beta_{pg}} & \sim \norm(m_{\beta_{pg}}, s^2_{\beta_{pg}}) .
\end{aligned}
\end{equation} 

We therefore define an interaction as significant if $0$ falls outside the posterior $2 \sigma$ interval of $m_{\beta_{pg}}$. In other words, the interaction between $p$ and $g$ is significant if

\begin{equation}
m_{\beta_{pg}} - 2 s_{\beta_{pg}} > 0
\end{equation}
\emph{or}
\begin{equation}
m_{\beta_{pg}} + 2 s_{\beta_{pg}} < 0
\end{equation}

Note that variational inference typically underestimates posterior variances \cite{blei2016variational} so such a designation of \emph{significant} will be under-conservative.

\section{Results}

\begin{figure}[h]
  \centering
  \includegraphics[width=.6\textwidth]{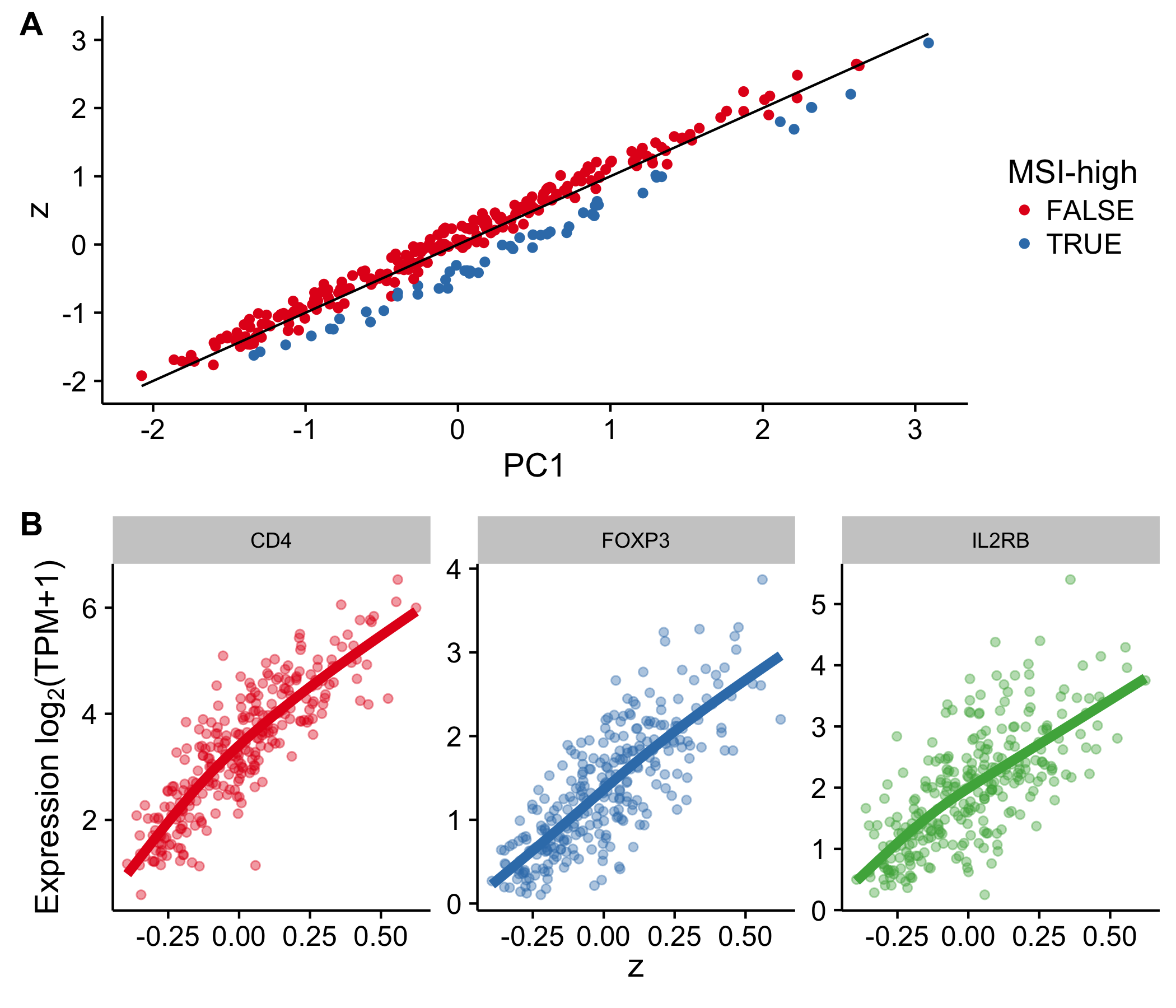}
  \caption{Test}
 \label{fig:trajectory}
\end{figure}

\begin{figure}[h]
  \centering
  \includegraphics[width=.5\textwidth]{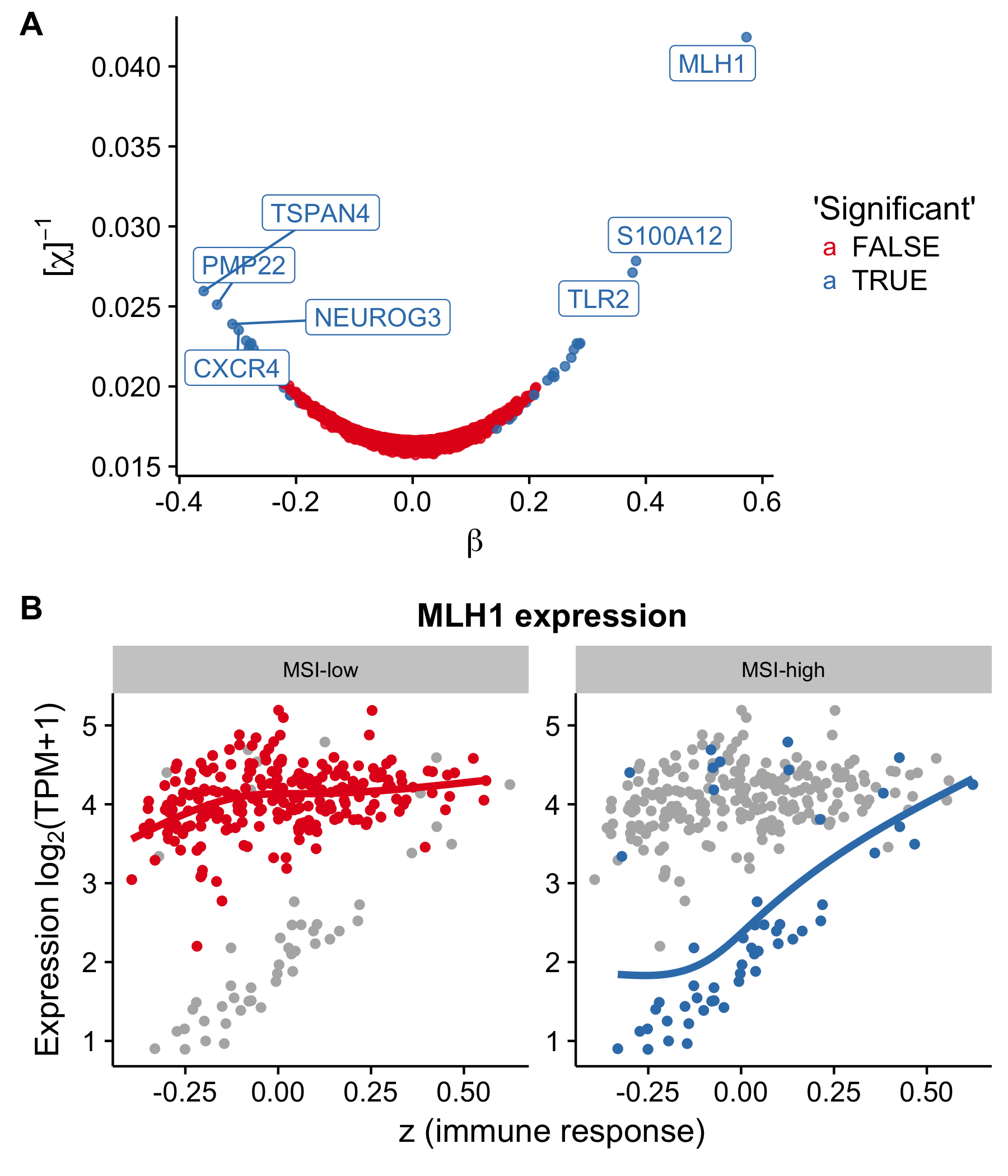}
  \caption{Test}
 \label{fig:beta}
\end{figure}

We applied our method to a recent quantification \cite{tatlow2016cloud} of the TCGA RNA-seq colorectal cancer dataset \cite{weinstein2013cancer}. After a quality-control step we retained 284 samples for analysis and used the 6394 genes whose variance in $\log_2(\text{TPM}+1)$ exceeded $0.5$. The covariate used for $\bX$ was whether the tumour exhibited high microsatellite instability (MSI), a hyper-mutable phenotype caused by the loss of DNA mismatch repair activity.  Tumours with MSI are known have a different response to chemotherapeutics along with marginally better prognosis \cite{boland2010microsatellite}.

We applied our covariate latent variable models to this data with strong shrinkage priors on $\beta$ ($a_\beta = 6$, $b_\beta=0.1$) on the assumption that such latent variable-covariate interactions are likely to be rare. Consequently, $\bz$ should largely recover the first principal component of the data, as seen in figure \ref{fig:trajectory}A. However, it can be seen when plotting $\bz$ against $\text{PC1}$ that the samples ``split'' based on MSI status. This is most likely due to the variation in gene expression caused by MSI status being absorbed by the $\alpha_{pg}$ coefficients of the model.

We next sought to calibrate our inferred trajectory with some external measure of progression. Survival analysis in TCGA data is problematic - measurements of survival are taken from initial prognosis with scarce recording of the assay timing relative to this. Furthermore, in the colorectal cancer dataset used 243 / 284 (85\%) patients have no survival information recorded at all. Consequently, we sought to compare our trajectory with a genomic measure of prognosis, namely \emph{FOXP3+} regulatory T-cell (Treg) status which is associated with poor colorectal cancer prognosis \cite{shang2015prognostic}. We examined the expression of three Treg markers along $\bz$ (\emph{CD25}, \emph{CD4} and \emph{FOXP3}, not included in $\bY$) which showed decrease in expression along $\bz$ implying a possible association between the inferred trajectory and prognostic potential.

Finally, we sought to identify significant interactions between genes, covariates and the trajectory. Using the significance criterion described in section \ref{sec:significant}, 35 genes were identified as showing possibly interesting interactions as can be seen on the $\beta$-$\chi^{-1}$ plot in figure \ref{fig:beta}A. Of the genes identified, \emph{MLH1} particularly stands out as having a strong association. \emph{MLH1} is known to be causal for MSI, either due to epigenetic silencing or a germline mutation \cite{vilar2010microsatellite}. To understand the precise (inter-)action of \emph{MLH1} we examined its behaviour along the immune-response trajectory coloured by MSI status (figure \ref{fig:beta}B). In the MSI-low regime the expression of \emph{MLH1} is constant independent of the immune ($\bz$) status. However, in the MSI-high regime the expression of \emph{MLH1} becomes dependent on the MSI-status of the tumour. As such, our covariate latent variable model has identified an interesting interaction between immune response, MSI-status and \emph{MLH1} expression.

% The results may be seen in Figures \ref{fig:coad}B-D. The strong sparsity priors on $\beta$ suggest the effect of the covariates $\bX$ on the inferred $\bt$ will be minimal and should be similar to traditional factor analysis models; however, Figure \ref{fig:coad}B clearly shows MSI status having a discernible shift on the inferred latent variables. While this effect is small in the dataset examined it is conceivable that larger confounding effects may exist in other datasets. We subsequently examined the posterior values of $\chi^{-1}$ and $\beta$ for each coefficient and gene to discover any interactions between the lipid-metabolic trajectory and the covariates included (Figure \ref{fig:coad}C). This identified genes such as \emph{NRF2}, \emph{ALDH3B1} and \emph{PPARD} all associated with colorectal cancer outcomes \cite{lee2005nrf2,khorrami2015verification,park2012role}.

\section{Discussion}

We have proposed the concept of replacing discrete disease staging with data-driven continuous trajectories inferred from genomics data that hold prognostic and/or diagnostic value. By considering a modified factor analysis model we incorporate population-level heterogeneity that may confound existing trajectory-based analysis. By applying our model to RNA-seq data from the TCGA colorectal cancer dataset we simultaneously identify genes that possibly interact with externally measured covariates while learning a trajectory that correlates with known markers of colorectal cancer prognosis.

One limitation of the model is its linear nature, making inferred latent variables similar to those from factor analysis. We therefore propose a nonlinear, nonparametric extension similar to Gaussian Process Latent Variable Models \cite{lawrence2004gaussian}. The trick is to consider the $\bX\bX^T$ term in the covariance matrix of the marginal distribution of $\bY$ and replace it with any (semi-)positive definite matrix representing ``similarity'' between the elements of $\by$, such as double-exponential kernels. We therefore mention the possibility of \emph{Covariate Gaussian Process Latent Variable Models} with kernels given by

\begin{equation}
K\left( \{\bx_{p = 1, \ldots, P}, \bz\}, \{\bx_{p = 1, \ldots, P}', \bz'\} \right) \propto K(\bx, \bx') + K(\bz, \bz') + 
\sum_p K(\bx_p \odot \bz, \bx_p' \odot \bz')
\end{equation}

for some suitable choice of kernel function $K$. Note that the element-wise product $\odot$ only appears because of the linear relationship between the covariates and factor loading matrix. This could easily be replaced by any nonlinear and possibly nonparametric function $\bff$ giving terms of the form $K\left( (\bff(\bx_p, \bz), \bff(\bx_p', \bz') \right)$.

Inference for linear C-LVM was possibly on a laptop for $\mathcal{O}(10^2)$ samples, $\mathcal{O}(10^4)$ features $\mathcal{O}(1)$  covariates. However, such inference would become burdensome for increasing data size, particularly with respect to features and covariates as $\mathcal{O}(PG)$ parameters are required.  Fortunately, the currently-implemented CAVI inference can be easily modified to give scalable Stochastic Variational Inference \cite{hoffman2013stochastic} for this model.

%\begin{itemize}
%\item Covariate GPLVM
%\item Variational inference because everything is conjugate
%\end{itemize}

{\small
\printbibliography 
}

\end{document}